\documentclass[letterpaper, 10 pt, conference]{ieeeconf}  
\IEEEoverridecommandlockouts   
\usepackage{floatrow}
\usepackage{afterpage}
\usepackage{cite}
\usepackage{algorithm} 
\usepackage{algorithmic}  
\usepackage[algo2e]{algorithm2e} 
\usepackage{amsmath,amssymb,amsfonts}
\usepackage{algorithm}
\usepackage{graphicx}
\usepackage{textcomp}
\def\BibTeX{{\rm B\kern-.05em{\sc i\kern-.025em b}\kern-.08em
    T\kern-.1667em\lower.7ex\hbox{E}\kern-.125emX}}
    
    \usepackage{lineno}
\usepackage{lettrine}

\usepackage{url}
\usepackage{moreverb}

\usepackage{graphicx}
\usepackage{subcaption}
\usepackage{float}
\usepackage{amsmath}
\usepackage{array}
\usepackage{multirow}
\graphicspath{ {Figures/} }

\usepackage{listings}
\usepackage{scalerel}
\usepackage{tikz}
\usetikzlibrary{svg.path}

\definecolor{orcidlogocol}{HTML}{A6CE39}
\tikzset{
  orcidlogo/.pic={
    \fill[orcidlogocol] svg{M256,128c0,70.7-57.3,128-128,128C57.3,256,0,198.7,0,128C0,57.3,57.3,0,128,0C198.7,0,256,57.3,256,128z};
    \fill[white] svg{M86.3,186.2H70.9V79.1h15.4v48.4V186.2z}
                 svg{M108.9,79.1h41.6c39.6,0,57,28.3,57,53.6c0,27.5-21.5,53.6-56.8,53.6h-41.8V79.1z M124.3,172.4h24.5c34.9,0,42.9-26.5,42.9-39.7c0-21.5-13.7-39.7-43.7-39.7h-23.7V172.4z}
                 svg{M88.7,56.8c0,5.5-4.5,10.1-10.1,10.1c-5.6,0-10.1-4.6-10.1-10.1c0-5.6,4.5-10.1,10.1-10.1C84.2,46.7,88.7,51.3,88.7,56.8z};
  }
}

\newcommand\orcidicon[1]{\href{https://orcid.org/#1}{\mbox{\scalerel*{
\begin{tikzpicture}[yscale=-1,transform shape]
\pic{orcidlogo};
\end{tikzpicture}
}{|}}}}

\usepackage{hyperref} %<--- Load after everything else

\begin{document}
\title{\LARGE \bf
LLM-Guided Transformation of Non-Critical Driving Scenes into Safety-Critical Scenarios Using Augmented Reality}
    
\author{Noura Fady$^{1,2}$,  Farah Khaled$^{1,2\orcidicon{0009-0001-6869-1288}\,}$, and Catherine~M.~Elias$^{1,2\orcidicon{0000-0002-1444-9816}\,}$,~\IEEEmembership{Member,~IEEE,}% 
\thanks{*This work was not supported by any organization}% <-this % stops a space
\thanks{$^{1}$C-DRiVeS Lab: Cognitive Driving Research in Vehicular Systems, Cairo, Egypt
{\tt\small cdrives.researchlab@gmail.com}}%
\thanks{$^{2}$Computer Science and Engineering Department - Faculty of Media Engineering and Technology - German University in Cairo, Egypt}%
\thanks{{\tt\small efarahkhaled03@gmail.com, catherine.elias@ieee.org}}%
}

\markboth{Journal of \LaTeX\ Class Files,~Vol.~14, No.~8, August~2015}%
{author1 \MakeLowercase{\textit{et al.}}:title here}



\maketitle
\begin{abstract}
Testing Autonomous Driving Systems (ADS) requires realistic safety-critical scenarios, but collecting such data from real-world driving is costly and unsafe. This paper presents an automated pipeline that transforms safe driving scenes into safety-critical scenarios by combining computer vision, Large Language Models (LLMs), and Augmented Reality (AR). The system detects and tracks road users, extracts safety features including distance, velocity, motion direction, and Time-to-Collision (TTC), and assesses scene criticality. Safe scenes are modified by an LLM, which generates realistic collision-inducing objects and behaviors that are integrated into the original scene using AR. The proposed pipeline was evaluated on the nuScenes dataset, achieving 97.52\% safety classification accuracy and successfully generating realistic scenarios such as pedestrian crossings, rear overtaking vehicles, and sudden-stop events. The results demonstrate an effective and flexible approach for automated generation of safety-critical scenarios to support the testing and validation of autonomous driving systems. 
\end{abstract}


% \fi
%\IEEEpeerreviewmaketitle
\section{Introduction}\label{sec1}

Autonomous Driving Systems (ADS) enable vehicles to operate with minimal human intervention by combining artificial intelligence, onboard sensors, and modular perception, prediction, planning, and control components \cite{citeKey14,citeKey15,citeKey5}. Despite recent advances, ensuring safe and reliable operation remains a major challenge due to the complexity and unpredictability of real-world traffic \cite{citeKey1}. Consequently, extensive validation under diverse driving conditions is required before deploying ADS in real-world environments \cite{citeKey1,citeKey2,citeKey3,citeKey5,citeKey7}.

Real-world testing alone is insufficient because safety-critical events are rare, expensive, and dangerous to reproduce, limiting the coverage of critical edge cases \cite{citeKey1,citeKey3,citeKey7,citeKey15}. Therefore, scenario-based testing has emerged as an effective alternative for evaluating autonomous driving systems under controlled conditions \cite{citeKey1,citeKey2,citeKey3,citeKey7,citeKey5}. Recent progress in LLMs has further advanced scenario generation by enabling context-aware reasoning \cite{goba2025prompts, gado2026prompts, khairy2026cadenet} and automated generation of safety-critical situations, while Augmented Reality (AR) enables these generated modifications to be integrated into real driving scenes without compromising realism \cite{citeKey1}.

Although existing scenario-generation methods have improved autonomous driving validation, they still rely heavily on simulation environments and provide limited capability for generating realistic and diverse safety-critical scenarios directly from real-world driving scenes. Consequently, there remains a need for methods that systematically transform safe driving scenes into realistic safety-critical scenarios while preserving scene consistency. This work addresses this challenge by combining computer vision, LLM-guided scenario generation, and AR-based scene augmentation to automatically create realistic edge cases for autonomous driving testing.

\section{Literature Review}

Scenario-based testing has become the dominant approach for validating ADS, as real-world testing alone is costly, time-consuming, and insufficient in capturing rare safety-critical events \cite{citeKey6,citeKey8}. Traditional scenario-generation methods can be broadly categorized into data-driven, simulation-based, and knowledge-based approaches. While these methods enable controlled testing, they often struggle to generate diverse and realistic edge cases and remain highly dependent on predefined datasets or expert-designed rules \cite{citeKey6,citeKey3,citeKey10}.

Recent advances in LLMs have significantly improved scenario generation by enabling semantic reasoning, natural-language understanding, and structured scenario generation \cite{citeKey6,citeKey7}. LLMs have been applied to transform safe driving scenes into safety-critical situations, generate simulation scripts from textual descriptions, and support interactive retrieval-augmented frameworks for continuously expanding behavioral diversity \cite{citeKey1,citeKey2,citeKey10}. More recently, multimodal and foundation-model approaches have further enhanced scenario understanding and generation by integrating language with visual information, improving realism and scalability \cite{citeKey1,citeKey6}.

Despite these advances, existing methods remain largely dependent on simulation environments and rarely modify real driving scenes directly while preserving visual realism. Furthermore, many approaches provide limited control over generated scenarios and insufficient coverage of diverse real-world edge cases. These limitations motivate the proposed framework, which combines computer vision, LLM-guided scenario generation, and AR to transform real, non-critical driving scenes into realistic safety-critical scenarios.
\section{Methodology}\label{sec2}

\subsection{Overview of the Proposed Framework}

\begin{figure}[H]
    \centering
    \includegraphics[width=1\linewidth,
    trim={1cm 5cm 0cm 13cm},
    clip]
    {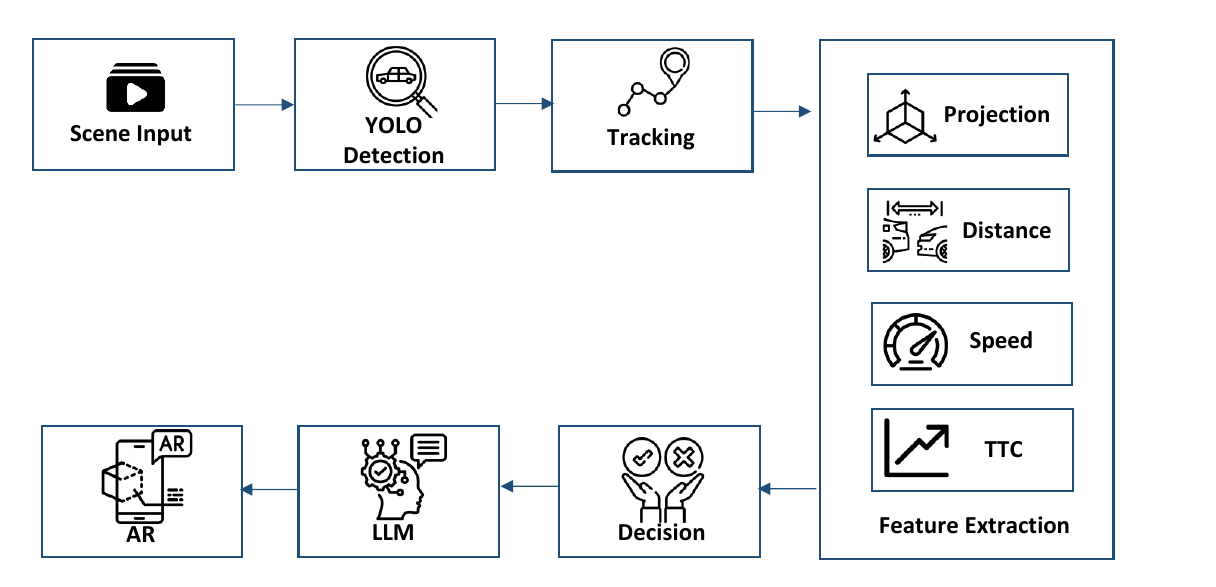}
    \caption{Overall pipeline of the proposed framework for transforming non-critical driving scenes into safety-critical scenarios.}
    \label{fig:pipeline}
\end{figure}

The proposed framework transforms non-critical driving scenes into safety-critical scenarios by integrating computer vision, safety assessment, LLMs, and AR, as illustrated in Fig. \ref{fig:pipeline}. 

The framework begins by detecting and tracking surrounding road users from a monocular driving video using YOLOv8 and centroid-based tracking. Projection-based geometry is then applied to estimate the spatial position of detected objects, allowing the computation of relative distance and velocity, which are used to calculate the Time to Collision (TTC). This is used to evaluate the safety level of the unmodified scene. When the scene is classified as safe, an LLM generates a structured description of a collision-inducing modification. Finally, the generated object is integrated into the original scene using AR techniques while preserving geometric consistency and visual realism.

\subsection{Scene Understanding and Safety Assessment}
% Each input frame is first processed using YOLOv8 to detect vehicles and pedestrians. Every detection provides a bounding box, object class, and confidence score. Consecutive detections are then associated through centroid-based tracking to maintain persistent identities throughout the video.

% Using the tracked object trajectories, three safety-related features are estimated:

% \begin{itemize}
%     \item Relative distance
%     \item Relative velocity
%     \item Time-to-Collision (TTC)
% \end{itemize}

% Relative distance is approximated using the pinhole camera model, where object size in the image is converted into metric distance based on camera calibration. Velocity is computed from frame-to-frame displacement of tracked objects, while TTC is calculated from the estimated distance and relative motion between the ego vehicle and surrounding agents. The smallest TTC among all detected objects represents the scene risk.

Each input frame is first processed using YOLOv8 to detect vehicles and pedestrians within a scene. For each detected object, the detector provides the object class, confidence score, and bounding-box coordinates. Consecutive detections are then associated through centroid-based tracking to maintain persistent object identities throughout different scenes.

\begin{figure}[h]
    \centering
    \includegraphics[width=0.95\linewidth]{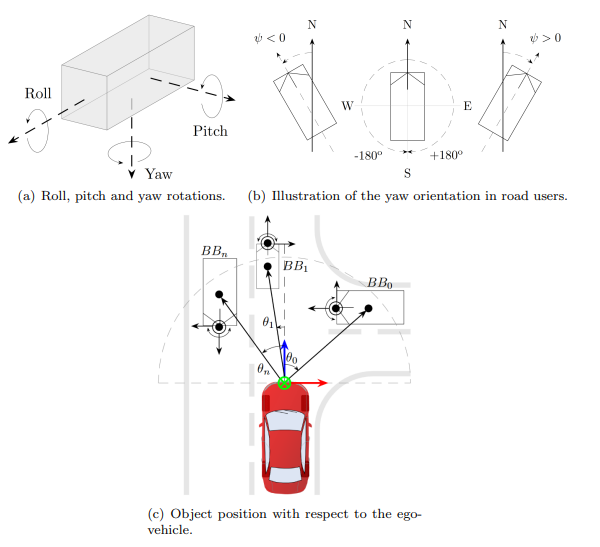}
    \caption{Estimation of the safety features used for scene evaluation.}
    \label{fig:image1}
\end{figure}

% \begin{equation}
% \mathrm{TTC} = \frac{d}{v_{\mathrm{rel}}}
% \label{eq:ttc}
% \end{equation}

% where d is the estimated relative distance and v
% rel is the relative velocity between the ego vehicle and the detected object.

As illustrated in Fig. \ref{fig:image1}, adapted from \cite{citeKey14}, the tracked object trajectories are used to estimate three safety-related features: relative distance, relative velocity, and TTC. Relative distance, as shown in equation \ref{eq:distance}, is estimated using the pinhole camera model, where the detected object size in the image is converted into metric distance based on the camera calibration parameters. 
% Object velocity is subsequently computed from the displacement of tracked objects across consecutive frames, as demonstrated in equation \ref{eq:velocity}, while TTC is calculated from the estimated relative distance and relative motion between the ego vehicle and surrounding road users. The minimum TTC among all tracked objects is selected as the representative safety measure of the scene and is used to determine its criticality.

\begin{equation}
d=\frac{fH}{h}
\label{eq:distance}
\end{equation}

where $f$ is the camera focal length, $H$ is the real object height, and $h$ is the detected object height in pixels.
\newline

Object velocity is subsequently computed from the displacement of tracked objects across consecutive frames, as demonstrated in equation \ref{eq:velocity}.

\begin{equation}
v=\frac{\sqrt{(\Delta x)^2+(\Delta z)^2}}{\Delta t}
\label{eq:velocity}
\end{equation}

where $\Delta x$ and $\Delta z$ denote the displacement between two consecutive frames, and $\Delta t$ is the elapsed time.
 \newline
 
Finally, TTC, shown in equation \ref{eq:ttc}, is calculated from the estimated relative distance and relative motion between the ego vehicle and surrounding road users.

\begin{equation}
\mathrm{TTC}=
\frac{d_{obj,ego}}
{v_{ego}-v_{obj}\cos(\psi_{obj})}
\label{eq:ttc}
\end{equation}

where $d$ is the estimated relative distance between the ego vehicle and the detected object, and $v_{\mathrm{rel}}$ is their relative velocity. 
\newline

The minimum TTC among all tracked objects is selected as the representative safety measure of the scene. If the computed TTC is below the predefined threshold of 1.5\,s, the scene is classified as safety-critical; otherwise, it proceeds to the LLM-based scenario generation stage.

\subsection{Safety-Critical Scenario Generation}

Whenever the TTC exceeds the predefined safety threshold, the scene is considered non-critical and is forwarded to the LLM. The prompt contains semantic information describing the current scene, composed of the scene type, detected objects, relative distances, and computed TTC values. Based on this context, the LLM generates a structured JSON response describing a new object capable of increasing collision risk.

% The generated response specifies the object type, initial position, relative location, and motion parameters while following predefined constraints to ensure realistic behavior. This semantic description forms the basis for the subsequent AR augmentation stage.

\begin{lstlisting}[caption=LLM prompt structure and generated scenario description., label = fig:llm_prompt, linewidth=\columnwidth,breaklines=true, frame=single]
    "Role: Autonomous Driving Scenario Generator
    
    Environment: University Campus
    
    Rules:
    - Follow road semantics.
    - Generate only realistic traffic events.
    - Do not generate pedestrians without sidewalks.
    - Do not generate crossing vehicles without intersections.
    - Return only valid JSON output.
    
    Input:
    TTC = 2.1 s 
    Detected Objects = Vehicle, Pedestrian 
    Road Type = Intersection
    
    Task:
    Generate a safety-critical scenario while
    maintaining environmental consistency."
\end{lstlisting}

As shown in listing \ref{fig:llm_prompt}, the LLM generates a structured JSON response describing a new collision-inducing object. The generated output specifies the object type, initial distance, speed, and relative position while maintaining consistency within the surrounding environment. This structured representation serves as the bridge that closes the gap between semantic reasoning and AR-based scene generation.

\subsection{Augmented Reality Scene Generation}

\begin{figure}[H]
    \centering
    \includegraphics[width=0.9\linewidth]{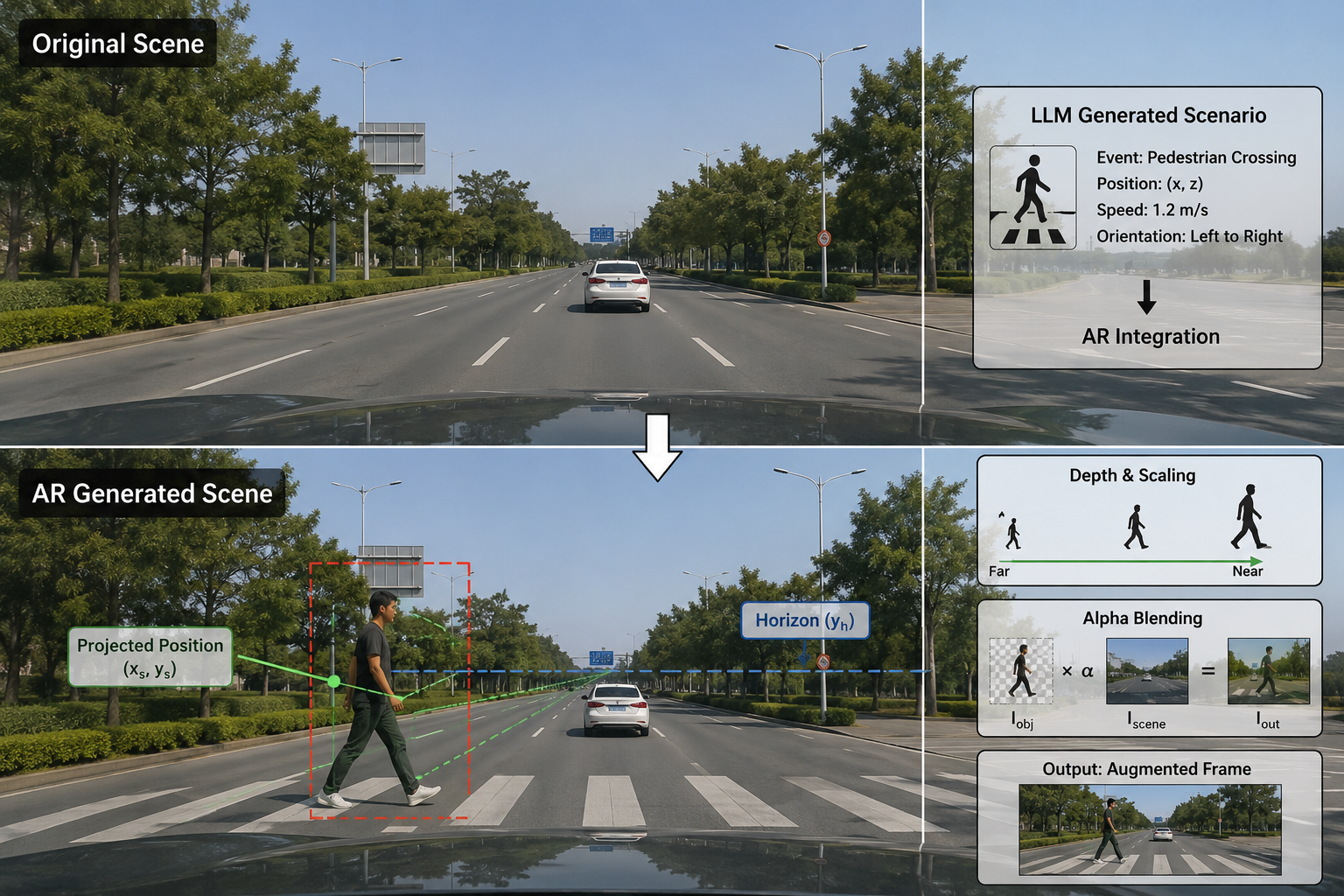}
    \caption{AR techniques used to preserve geometric and temporal consistency}
    \label{fig:image2}
\end{figure}

The generated semantic description is converted into a visual object that is inserted into the original driving video. To preserve realism, three complementary techniques are employed, as illustrated in Fig. \ref{fig:image2}.

First, depth-aware scaling adjusts the object size according to its estimated distance from the camera. Second, perspective projection aligns the generated object with the road geometry using the camera model. Finally, optical-flow-based motion compensation accounts for ego vehicle movement to ensure temporal consistency across consecutive frames.

These operations produce realistic safety-critical scenes while preserving the appearance and context of the original recording.

\subsection{Implementation Details}

\begin{table}[h]
\caption{Implementation parameters of the proposed framework.}
\label{tab:implementation}
\centering
\begin{tabular}{ll}
\hline
\textbf{Parameter} & \textbf{Value} \\
\hline
Object Detector & YOLOv8s \\
Tracking Algorithm& Centroid Tracker \\
Input Resolution & $1280 \times 720$ \\
LLM Model& Llama 3 (Ollama) \\
TTC Threshold & 1.5 s \\
Programming Language & Python \\
AR Engine & OpenCV \\
Dataset & nuScenes \\
\hline
\end{tabular}
\end{table}

\section{Results \& Discussion}\label{sec4}

The proposed framework was evaluated using the nuScenes dataset, where predicted scene classifications were compared against the corresponding ground-truth safety labels. The evaluation was performed on 10 driving scenes comprising 404 video frames, each classified as either Safe or Unsafe according to the computed TTC. The generated classifications were then compared with those derived from the nuScenes annotations to assess the reliability of the proposed safety assessment module.

The quantitative evaluation demonstrated an overall classification accuracy of 97.52\%, with 394 out of 404 frames correctly , as shown in table \ref{tab:nuScenes_results}. Most evaluated scenes achieved 100\% classification accuracy, while the remaining scenes maintained accuracies above 89\%, demonstrating the robustness of the proposed safety assessment framework across different driving conditions.

\begin{table}[h]
\centering
\footnotesize
\caption{Comparison between the proposed pipeline and nuScenes safety classifications.}
\label{tab:nuScenes_results}

\begin{tabular}{|c|c|c|c|c|}
\hline
\textbf{Scene} &
\textbf{Total Frames} &
\textbf{Correct} &
\textbf{Incorrect} &
\textbf{Accuracy (\%)} \\
\hline

1 & 41 & 41 & 0 & 100.00 \\
2 & 39 & 35 & 4 & 89.74 \\
3 & 40 & 40 & 0 & 100.00 \\
4 & 41 & 41 & 0 & 100.00 \\
5 & 41 & 41 & 0 & 100.00 \\
6 & 41 & 41 & 0 & 100.00 \\
7 & 40 & 40 & 0 & 100.00 \\
8 & 41 & 38 & 3 & 92.68 \\
9 & 40 & 40 & 0 & 100.00 \\
10 & 40 & 37 & 3 & 92.50 \\
\hline
\textbf{Overall} & \textbf{404} & \textbf{394} & \textbf{10} & \textbf{97.52} \\
\hline

\end{tabular}
\end{table}

\begin{figure}[h]
    \centering
    \includegraphics[width=0.95\linewidth]{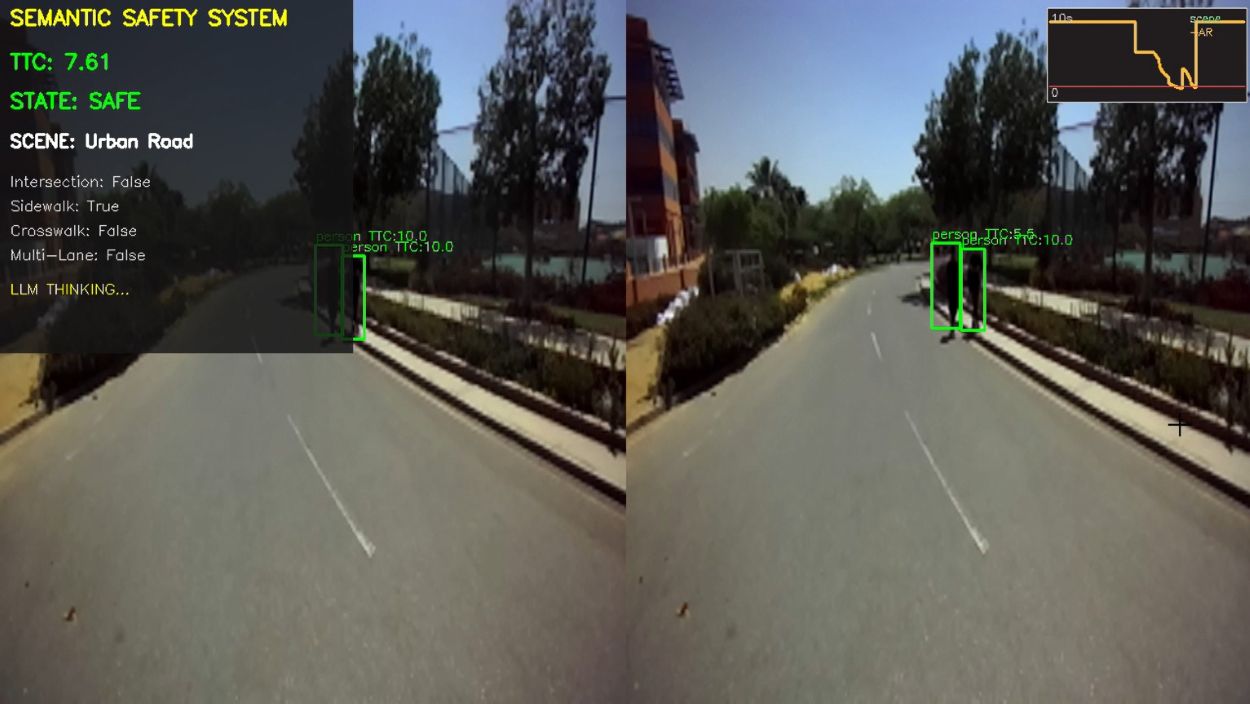}
    \caption{ Original scene before AR augmentation.}
    \label{fig:before}
\end{figure}

\begin{figure}[h]
    \centering
    \includegraphics[width=0.95\linewidth]{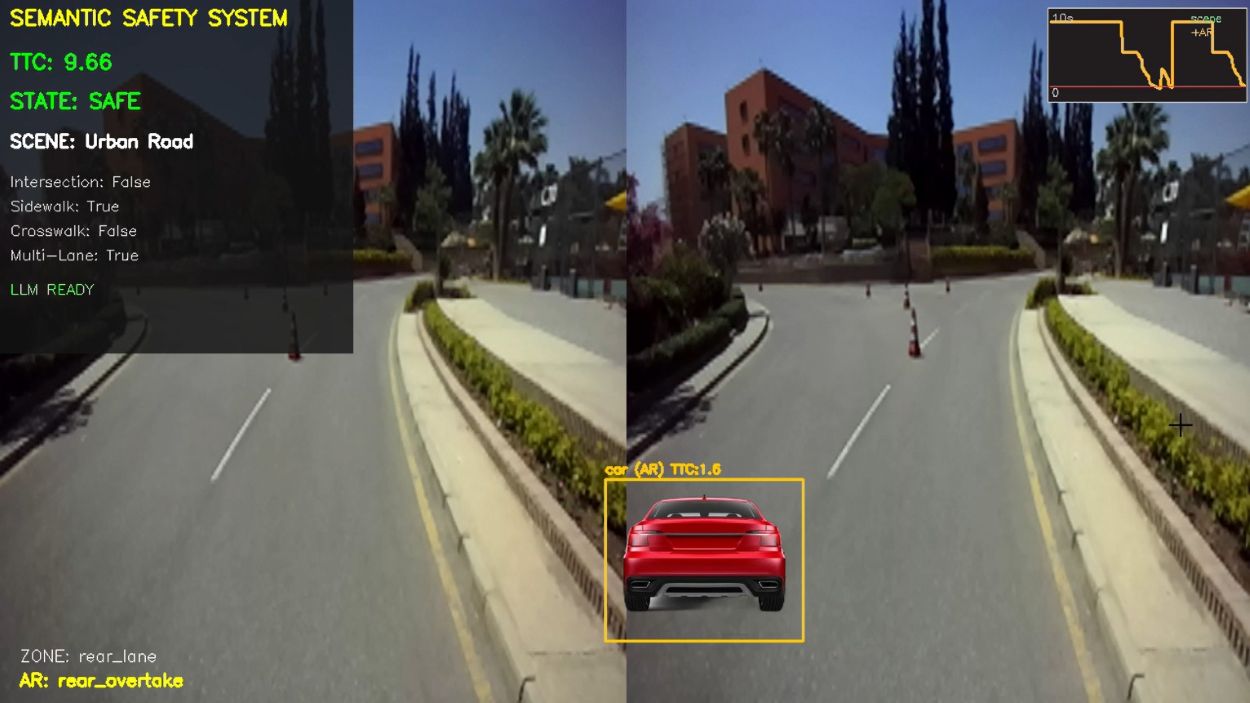}
    \caption{Safety-critical scene after AR augmentation.}
    \label{fig:after}
\end{figure}

Beyond quantitative evaluation, the proposed framework successfully generated multiple realistic safety-critical scenarios from originally safe driving scenes. The LLM produced context-aware modifications describing new collision-inducing objects, while the AR module seamlessly integrated these objects into the original video, even in real-time scenarios as demonstrated by Fig. \ref{fig:before} and Fig. \ref{fig:after}, preserving spatial consistency and scene realism.

Finally, qualitative observations showed that the generated scenarios preserved realistic object placement, perspective, and motion while introducing meaningful collision risks. These results demonstrate that the proposed framework can automatically transform non-critical driving scenes into realistic safety-critical scenarios suitable for autonomous driving testing and validation.
\section{Conclusion and Future Recommendations}\label{sec4}

This paper presented an automated framework for transforming non-critical driving scenes into realistic safety-critical scenarios by integrating computer vision, LLMs, and AR. The proposed pipeline performs object detection, tracking, and TTC analysis to assess scene safety before generating context-aware hazard modifications through an LLM. These generated objects are then integrated into the original driving scene using AR while preserving geometric consistency and visual realism. Experimental evaluation on the nuScenes dataset demonstrated the reliability of the proposed safety assessment framework, achieving an overall classification accuracy of 97.52\% and successfully generating diverse safety-critical scenarios suitable for autonomous driving testing and validation. Furthermore, limited real-time testing aided in validating the integrity of the pipeline and highlighted its ability for further use in more realistic scenarios beyond datasets.

Future work will focus on improving object tracking and trajectory prediction, incorporating richer scene understanding through additional sensors and multi-modal perception, and extending the framework to more complex traffic environments. Furthermore, integrating more advanced AR rendering techniques and conducting large-scale real-world validation will improve the realism, scalability, and practical applicability of the proposed framework for autonomous driving evaluation.
\appendices

\bibliographystyle{IEEEtran}
\bibliography{sections/ref} %edit all the generated bibtex from google scholar in sections/ref.bib

\end{document}